\documentclass[11pt,a4paper]{article}
\usepackage[hyperref]{acl2019}
\usepackage{times}
\usepackage{latexsym}
\usepackage{url}
\usepackage{booktabs}       
\usepackage{amsfonts}       
\usepackage{nicefrac}       
\usepackage{microtype}      
\usepackage{amssymb}
\usepackage{multirow}
\usepackage{multicol}
\usepackage{graphicx}
\usepackage{courier}
\usepackage{wrapfig}
\usepackage{adjustbox}
\usepackage{tabularx}
\usepackage{diagbox}
\usepackage{makecell}
\usepackage{caption}
\usepackage{enumitem}
\usepackage{titlesec}
\usepackage{natbib}
\usepackage{caption}
\usepackage{soul}
\usepackage{enumitem}
\titlespacing{\section}{0pt}{*1.0}{*1.0}
\usepackage[hang]{footmisc}

\usepackage{letltxmacro}

\aclfinalcopy 

\title{Domain-Independent turn-level Dialogue Quality Evaluation via User Satisfaction Estimation}

\author{\\Praveen Kumar Bodigutla, \space Longshaokan Wang, Kate Ridgeway, Joshua Levy, \\\texttt{\{pbodigut, longsha, kspridge, levyjos\}@amazon.com},\\Swanand Joshi\thanks{Currently at Facebook, but did this work at Amazon.}\space, Alborz Geramifard\footnotemark [1]\space, Spyros Matsoukas \\\texttt{\{swanandj7, alborzgeramifard\}@gmail.com, matsouka@amazon.com} \\\\Amazon Inc, USA }

\begin{document}
\maketitle

\begin{abstract}
An automated metric to evaluate dialogue quality is vital for optimizing data driven dialogue management. The common approach of relying on explicit user feedback during a conversation is intrusive and sparse. Current models to estimate user satisfaction use limited feature sets and rely on annotation schemes with low inter-rater reliability, limiting generalizability to conversations spanning multiple domains. To address these gaps, we created a new {\em Response Quality} annotation scheme, based on which we developed turn-level  {\em User Satisfaction} metric. We introduced five new domain-independent feature sets and experimented with six machine learning models to estimate the new satisfaction metric.

Using Response Quality annotation scheme, across randomly sampled single and multi-turn conversations from $26$ domains, we achieved high inter-annotator agreement (Spearman's rho $0.94$). The Response Quality labels were highly correlated ($0.76$) with explicit turn-level user ratings. Gradient boosting regression achieved best correlation of {\scriptsize $\sim$}$0.79$ between predicted and annotated user satisfaction labels. Multi Layer Perceptron and Gradient Boosting regression models generalized to an unseen domain better (linear correlation $0.67$) than other models.  Finally, our ablation study verified that our novel features significantly improved model performance.

\end{abstract}
\section{Introduction}

\label{sec:intro}
Automatic turn and dialogue level quality evaluation of end user interactions with spoken dialogue systems is vital for identifying problematic conversations and for optimizing dialogue policy using a data driven approach, such as reinforcement learning. One of the main obstacles to designing data-driven policies is the lack of an objective function to measure the success of a particular interaction. Existing methods along with their limitations to measure dialogue success can be categorized into 4 groups: 1) Using sparse sentiment for end-to-end dialogue system training; 2) Using task success as dialogue evaluation criteria, which does not capture frustration caused in intermediate turns and assumes end user goal is known in advance;  3) Explicitly soliciting feedback from the user, which is intrusive and causes dissatisfaction; and 4) Estimating per turn dialogue quality using trained Interaction Quality (IQ) \cite{SCHMITT12.333} estimation models. Per turn discrete 1-5 scale IQ labels are provided by the annotator while keeping track of dialogue quality till that turn. This approach to rate turns increases cognitive load on the annotators and makes it difficult to pin-point defective turns. IQ annotations are reliable in comparison to explicit satisfaction ratings provided by user at the end of the dialogue (an approach followed by PARADISE \cite{Walker:2000:TDG:973935.973945} dialogue evaluation framework). However, IQ annotation scheme which is developed using within domain conversations has limited generalizability to dialogues which span multiple domains.

Various models have been explored to predict IQ, including Hidden Markov Models \cite{10.1007/978-1-4614-8280-2_27}, Support Vector Machines (SVM) \cite{Schmitt2011ModelingAP}, Support Vector Ordinal Regression (SVOR) \cite{6854195}, Recurrent Neural Networks (RNN) \cite{Pragst2017}, and most recently, Long Short-Term Memory Networks (LSTM) \cite{Rach2017InteractionQE}. Features used in these models were derived from the current turn, the dialogue history, and output from three Spoken Language Understanding (SLU) components, namely: Automatic Speech Recognition (ASR), Natural Language Understanding (NLU), and the dialogue manager. We hypothesize that these features are limited, and including additional contextual signals improves the performance of dialogue quality estimation models for both single-turn and multi-turn conversations\footnote{In single-turn conversations the entire context is expected to be present in the same turn, while in a multi-turn case the context is carried from previous turns to address user's request in the current turn.} (example in Appendix Table \ref{multi-domain-example}). To this end we designed new features to capture the following: user rephrase, cohesion between user request and system response, diversity of topics discussed, un-actionable user requests, and popularity of domains and topics across the entire population of the users.

To obtain consistent, simple and generalizable annotation scheme which easily scales to multi-domain conversations, we introduce Response Quality (RQ) annotation scheme. We propose an end-to-end User Satisfaction Estimation (USE) metric which predicts turn level user satisfaction rating on a continuous 1-5 scale. Using RQ annotation scheme, annotators rated dialogue turns from $26$ single-turn and multi-turn sampled Alexa (commercial speech-based assistant) domains (e.g., \textit{Music, Calendar, Weather, Movie booking}). With the help of the new features we introduced, we trained four USE machine learning models using the annotated RQ ratings. We reserved one ``new'' multi-turn Alexa skill \cite{DBLP:journals/corr/abs-1711-00549}, to test USE models' performance on data from a new unseen domain. To explain model predicted rating using features' values, we experimented with interpretable models that rank features by their importance. We benchmarked performance of these models against two state-of-the-art dialogue quality prediction models. Using ablation studies, we showed improvement in the best performing USE model's performance using new contextual features we introduced. 

The outline of the paper is as follows: Section 2 introduces the Response Quality annotation scheme and discusses its effectiveness in terms of predicting user satisfaction ratings. Section 3 summarizes the Response Quality annotated data and our experimentation setup. Section 4 provides results from feature ablation study and an empirical study of six machine learning models to predict user satisfaction ratings on seen and unseen domains. Section 5 concludes.

\section{Response Quality Annotation}
We designed the Response Quality (RQ) annotation scheme to generate training data for our User Satisfaction Estimation (USE) model. 
\label{sec:annotation}
\subsection{RQ annotation Scheme and Comparison with IQ}
In RQ, similar to IQ annotations, annotators listened to raw audio and provided per turn's system RQ rating on an objective 5-point scale.  The scale we asked annotators to follow was: $1$=Terrible (fails to understand user's goal), $2$=Bad (understands goal but fails to satisfy it in any way), $3$=OK (partially satisfies the goal), $4$=Good (mostly satisfies the goal), and $5$=Excellent (completely satisfies the user's goal). 

Annotators rated conversations which spanned multiple domains and skills. They were instructed to use the follow-up feedback from the user (e.g., user expresses frustration or rephrases an initial request) in making judgements. Unlike IQ annotation scheme, we removed the constraint on the annotators to keep track of the quality of dialogue so far while determining RQ ratings for a given turn. This relaxation in constraint, coupled with making full conversation context available to the annotators, reduced the cognitive load on them. This simplified annotation scheme not only helped in scaling RQ to multiple domains and skills but also enabled precise identification of defective turns which is not straightforward in the case of IQ where an individual turn's IQ rating depends on the prior turns' ratings.

\subsection{Inter Annotator Agreement (IAA) and Correlation with user satisfaction rating \label{sec:IAA}}

We conducted a user study to verify the accuracy of RQ and IQ \cite{SCHMITT12.333} annotation process. In the study, eight users were asked to achieve $30$ pre-determined goals which were sampled from six single-turn and two multi-turn Alexa domains. For $15$ out of the $30$ goals, we asked the users to provide satisfaction rating on a discrete (1-5) scale based on turn's Alexa response. For the remaining $15$ goals, we asked the users to rate each turn incrementally based on their perception of interaction so far. Then we sent the same utterances for RQ (950 turns) and IQ (700 turns) annotations. 

We found that the RQ ratings provided by $3$ annotators were highly correlated (Spearman's rho $0.94$) with each other, suggesting high IAA. The mean RQ ratings were significantly (at 95\% confidence interval) correlated ($0.76$) with surveyed user satisfaction ratings with Alexa's response. In the case of IQ ratings, IAA and correlation with user ratings (based on dialogue so far) dropped to $0.26$ and $0.36$ respectively, suggesting limited generalizability of IQ annotation scheme to multi-domain conversations.

\section{Data and Experimental Setup} 
This section describes our dataset, details the list of features derived from various signals, and explains our experimentation setup.
\vspace{-0.1cm}
\subsection{Data}
\begin{figure}[!h]
\captionsetup{font=small}
  \begin{center}
    \includegraphics[width=0.5\textwidth]{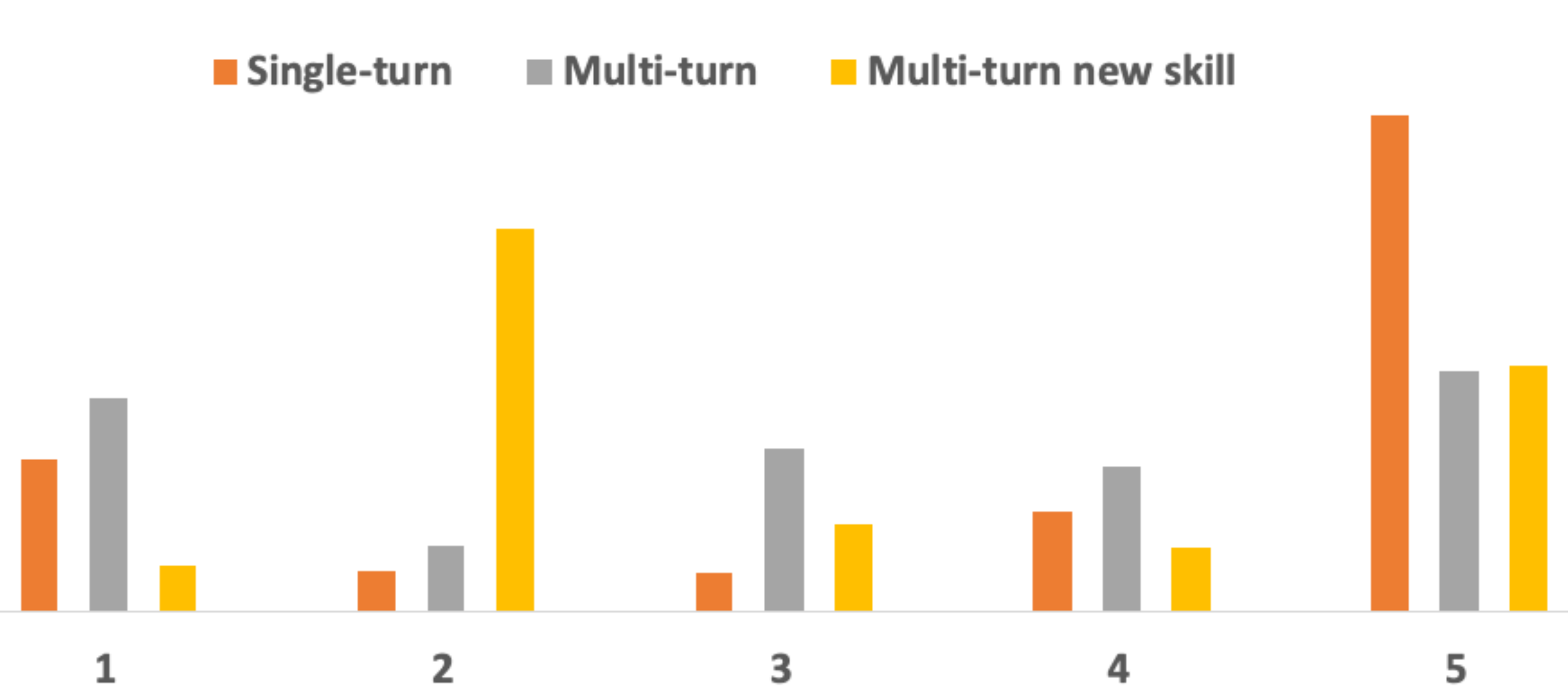}
  \end{center}
  \caption{Distribution of RQ annotation for single-turn, multi-turn domains and new multi-turn skill. Exact percentage on y-axis is masked for confidentiality}
  \label{fig:rating-distribution}
\vspace{-4mm}
\end{figure}

\vspace{-0.1cm}
\label{subsec:data}
To demonstrate that our RQ annotation scheme and predictive models are domain-independent and effective for both single-turn and multi-turn dialogues, we used $30,500$ dialogue turns randomly sampled from $26$ single-turn ($90\%$) and multi-turn ($10\%$) sampled Alexa domains. The imbalance towards single-turn dialogues is due to annotation priority. We also tested model's generalization performance on $200$ dialogue turns sampled from a ``new'' multi-turn goal oriented skill. Figure \ref{fig:rating-distribution} shows the diversity in rating distribution between the single-turn and multi-turn dialogues.

\label{subsec:feat}

\label{subsec:feat}
\begin{figure}[!h]
\captionsetup{font=small}
  \begin{center}
    \includegraphics[width=0.5\textwidth]{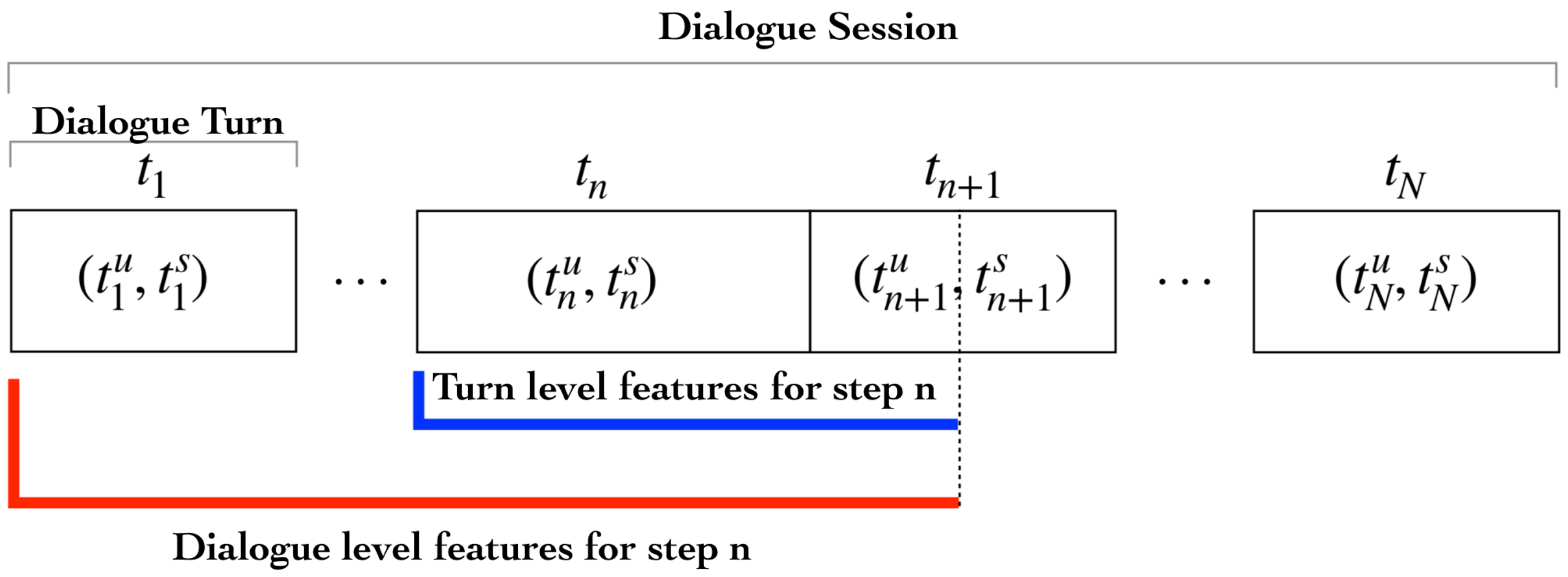}
  \end{center}
  \caption{dialogue and turn definitions. The blue and red lines indicate the history used for generating turn-based and dialogue-based features for the user satisfaction estimation on turn $t_n$.}
  \label{fig:turn-dialog-parameters}
\vspace{-6mm}
\end{figure}

\subsection{Features}
To estimate the turn level user satisfaction score, we used features derived from turn, dialogue context, and Spoken Language Understanding (SLU) components' output similar to turn level IQ prediction models \cite{Ultes2017DomainIndependentUS}. As shown in Figure \ref{fig:turn-dialog-parameters}, we define a dialogue turn at time $n$ as $t_n=(t_n^u, t_n^s)$, where $t_n^u$ and $t_n^s$ represent the user request and system response on turn $n$. A dialogue session of $N$ turns is defined as ($t_1$:$t_N$).  To improve the performance of the USE models across single and multi-turn conversations spanning multiple domains, we introduced the following $5$ sets of domain-independent features: 

\begin{enumerate}[nolistsep]
\item \textbf{User request paraphrasing} -- Calculated by measuring syntactic and semantic (NLU predicted intent) similarity between consecutive turns' user utterances. 

\item \textbf{Cohesion between response and request} -- Cohesiveness of system response with a user request is computed by calculating jaccard similarity score between user request and system response. System response \textit{``Here is a sci-fi movie''} will get higher cohesion score over \textit{``Here is a comedy movie''}, if the user request was \textit{``recommend a sci-fi movie''}.

\item \textbf{Aggregate topic popularity} -- Usage statistics such as aggregate domain and intent usage count and ratio of usage count to number of customers, provides us a prior on the popularity of a topic across all users of Alexa.

\item \textbf{Un-actionable user request} -- Identifies if the user request could not be fulfilled, by searching for phrases indicating an apology and negation in system response (e.g., \textit{``sorry I don't know how to do that''}). 

 \item \textbf{Diversity of topics in a session} --  This dialogue level feature is calculated using the percentage of unique intents till the current dialogue turn.
\end{enumerate}

\subsection{Experimental Setup}
\label{subsec:setup}
We obtained a turn's RQ rating by averaging the discrete 1-5 labels provided by $3$ annotators. Hence, we considered regression models for experimentation which predicted satisfaction rating on a continuous 1-5 scale.  To achieve interpretability, in our experiments we selected four models - LASSO , Decision Tree Regression,  Random Forest Regression and Gradient Boosting Regression that rank features by importance. For benchmarking we used Multi-layer Perceptron (MLP) and Support Vector Regression (SVR) \footnote{Recurrent Neural Networks and Long Short-term Memory Networks \cite{Rach2017InteractionQE} showed non-significant improvement ($\leq0.02$ difference in Spearman's rho) over state-of-the-practice MLP and SVM models in predicting IQ.}.

\subsubsection{Evaluation Criteria}
We used Pearson's linear correlation coefficient ($r$) for evaluating each model's 1-5 prediction performance.  For the use case to identify problematic turns from an end user's perspective, it is sufficient to identify {\em satisfactory} (rating $\geq$3) and {\em dissatisfactory} (rating $<3$) interactions. We used F-score for the dissatisfactory class\footnote{Notice that identifying dissatisfactory turns is of more importance and is in general a difficult task as majority of turns belong to the satisfactory class (Figure \ref{fig:rating-distribution}).} as the binary classification metric.

\section{Results and Analysis}
Amongst the six models we experimented with, Gradient Boosting Regression achieves superior performance on single-turn domains and multi-turn skill ($r$ = $0.79$ and F-dissatisfaction = $0.77$). On the new domain, both Gradient Boosting Regression and MLP models achieve better performance ($r$ = $0.67$ and F-dissatisfaction = $0.79$) in comparison to the other four models we experimented with \footnote{Results obtained on all $6$ models and optimal model parameters in Appendix Tables \ref{results} \& \ref{optimal-hyper-parameter}.}.

Based on ablation study using Gradient boosting regression model, we found that the new features improved every single metric on the test set. On single-turn dialogues, features corresponding to ``aggregate topic popularity'' caused largest statistically significant {\scriptsize $\sim$}$7\%$ relative improvement in linear correlation (0.741 $\rightarrow$ 0.796) and F-dissatisfaction (0.72 $\rightarrow$ 0.77) scores. On the new skill, {\scriptsize $\sim$}$35\%$ relative improvement in linear correlation (0.496 $\rightarrow$ 0.67) shows significant impact of ``Un-actionable request'' feature on generalization performance. The five new feature sets we introduced occur in the top 10 sets of important features (Appendix Table \ref{top-feature-weights}) returned by model based on their computed importance score.

\section{Conclusion}
In this paper, we described a user-centric and domain-independent approach for evaluating user satisfaction in dialogues with an AI assistant. We introduced Response Quality (RQ) annotation scheme which is highly correlated ($r$ = $0.76$) with explicit turn level user satisfaction ratings. By designing five additional new features, we achieved a high linear correlation of {\scriptsize $\sim$}0.79 between annotated RQ and predicted User Satisfaction ratings with Gradient Boosting Regression as the User Satisfaction Estimation (USE) model, for both single-turn and multi-turn dialogues. Gradient Boosting Regression and Multi Layer Perceptron (MLP) models generalized to unseen domain better ($r$ = $0.67$) than other models. 
 
With statistically significant {\scriptsize $\sim$}$7\%$ and {\scriptsize $\sim$}$35\%$ relative improvement in linear correlation on existing domains and new multi-turn skill respectively, our ablation study supported our hypothesis that the new features improve model prediction performance. On multi-domain conversations, we plan to explore the use of Deep Neural Net models to reduce handcrafting of features \cite{Rach2017InteractionQE} and for jointly estimating end user satisfaction at turn and dialogue level, though these models reduce interpretability. To learn dialogue policies using reinforcement learning, we plan to experiment with proposed RQ based User Satisfaction metric as an alternative for reward modeling.

\section*{Acknowledgments}
We thank  RAMP and Intelligent Decisions teams at Amazon for helping with data annotations. 
\scriptsize
\bibliographystyle{acl_natbib}
\setlength{\bibsep}{0pt plus 0.1ex}
\bibliography{acl2019}
\nopagebreak[0]
\cleardoublepage
\appendix

\onecolumn
\section{Appendices}
\label{sec:appendix}
\begin{table*}[h!]
\captionsetup{font=small}
  \centering
  \resizebox{0.50\textheight}{!}{
  \begin{adjustbox}{max width=\textwidth}
  \bgroup
  \def\arraystretch{1.1}
  \begin{tabularx}{\linewidth}{lX}
    \toprule
    \multicolumn{1}{c}{\textbf{Speaker}}   & \multicolumn{1}{c}{\textbf{Utterance}} \\
    \midrule
    User \{\textit{Makes a popular request\}} &  \hspace{5mm}Play latest hits. \textit{\{domain:\textbf{Music}}\}\\
    System \{\textit{Addresses the request successfully\}} & \hspace{5mm}Shuffling from your playlist.\\
    User \{\textit{barges-in}\}   &\hspace{5mm}Stop. Cancel my evening appointment \\
    					      &\hspace{5mm}\textit{\{domain:\textbf{Calendar}}\}\\
    System \{\textit{identifies un-actionable user request}\} &\hspace{5mm}Sorry I don't know that one \\
    User \{\textit{Paraphrases}\} & \hspace{5mm}Cancel my 7pm event if it is raining  today \\
    						& \hspace{5mm}\textit{\{domain:\textbf{Calendar, Weather}\}}\\
    System \{\textit{Informs and asks a question}\} & \hspace{5mm}Today in Seattle, there is a 60\% chance of rain.\\
     &		\hspace{5mm}Should I cancel your 7pm event - Park Visit?\\
    User \{\textit{confirms}\} &\hspace{5mm}Yes please \textit{\{domain:\textbf{Global}\}} \\
    System \{\textit{executes the request}\} & \hspace{5mm}Alright, 7pm event - Park visit Canceled. \\
    \bottomrule
  \end{tabularx}
  \egroup
  \end{adjustbox}
  }
   \caption{Conversation spanning multiple domains.\protect\footnotemark[\value{footnote}]}
      \vspace{-2.5mm}
  \label{multi-domain-example}
\end{table*}
\footnotetext{Due to confidentiality consideration, this example is \textit{not} a real user conversation but authored to mimic the true dialogue.}

\begin{table}[h!]
\captionsetup{font=small}
\centering
\resizebox{0.50\textheight}{!}{%

\begin{tabular}{|c|c|c|c|c|c|c|}
\hline
Model\textbackslash Metric & $Cor_{s}$ & $F-dis_{s}$ & $Cor_{m.t}$ & $F-dis_{m.t}$ & $Cor_{n.s}$ & $F-dis_{n.s}$\\
\hline
Lasso & 0.702 $\pm$ 0.013 & 0.689 $\pm$ 0.020 & 0.681 $\pm$ 0.052 & 0.740 $\pm$ 0.058 & 0.620 $\pm$ 0.071 & 0.740 $\pm$ 0.058  \\
Decision Tree & 0.727 $\pm$ 0.013 & 0.702 $\pm$ 0.019 & 0.669 $\pm$ 0.053 & 0.713 $\pm$ 0.062 & 0.579 $\pm$ 0.076 & 0.734 $\pm$  0.055 \\
Random Forest &  0.766 $\pm$ 0.011 & 0.740 $\pm$ 0.019 & 0.736 $\pm$ 0.045 & 0.746 $\pm$ 0.054 & 0.610 $\pm$ 0.072 & 0.763 $\pm$  0.051 \\
G.Boost & \textbf{0.795 $\pm$ 0.010} & \textbf{0.769 $\pm$ 0.017} & \textbf{0.792 $\pm$ 0.037} & \textbf{0.776 $\pm$ 0.047} & 0.670 $\pm$ 0.064 & \textbf{0.799 $\pm$  0.050} \\
SVR & 0.732 $\pm$ 0.012 & 0.715 $\pm$ 0.019 & 0.753 $\pm$ 0.042 & 0.710 $\pm$ 0.059 & 0.641 $\pm$ 0.068 & 0.764 $\pm$  0.053 \\
MLP & 0.745 $\pm$ 0.012 & 0.731 $\pm$ 0.020 & 0.739 $\pm$ 0.044 & 0.730 $\pm$ 0.055 & \textbf{0.671 $\pm$ 0.063} & 0.785 $\pm$ 0.050 \\
\hline
\end{tabular}
}
\vspace{-0.0cm}
\caption{Six machine learning regression models performance on turns from single-turn (subscript - s), multi-turn (subscript - m.t) conversations and new multi turn test skill (subscript - n.s). Each cell shows the mean and 95\% confidence interval with the highest mean in bold. Performance is measured using Linear correlation ($Cor$) and F-dissatisfaction (F-dis) scores.}
\vspace{-2.5mm}
  \label{results}
\end{table}

\begin{table*}[h!]
\captionsetup{font=small}
\centering
    \resizebox{0.50\textheight}{!}{
      \bgroup
  \begin{adjustbox}{max width=\textwidth}
  \def\arraystretch{1.1}
  \begin{tabularx}{\textwidth}{X X}
    \toprule
    \multicolumn{1}{c}{\textbf{Model}}  & \multicolumn{1}{c}{\textbf{Hyper parameter - Optimal value}}  \\
    \midrule
   Lasso & alpha:\,$0.001$\\
   Decision Trees &  max-depth:\,$33$,\,min-samples-leaf:\,$31$,\,min-samples-split:\,$23$ \\
   Random Forest &  max-depth:\,$49$,\,min-samples-leaf:\,$11$,\,min-samples-split:\,$27$ \\
   Gradient Boosting Decision Trees &  max-depth:\,$23$,\,min-samples-leaf:\,$17$,\,min-samples-split:\,$59$ \\
   SVR & c:\,$2$,\,gamma:\,$0.024$\\
   MLP & n-layers:\,$3$,\,batch-size:\,$128$,\,hidden size\,:\,$100$,\,solver:\,`sgd'\,activation:\,`relu'\\  
  \bottomrule
  \end{tabularx}
  \end{adjustbox}
    \egroup
  }
   \caption{Optimal Hyper parameter values used for training User Satisfaction Estimation (USE) models.}
  \label{optimal-hyper-parameter}
\end{table*}

\begin{table*}[h!]
\captionsetup{font=small}
  \centering
   \resizebox{0.50\textheight}{!}{
  \begin{adjustbox}{max width=\textwidth}
  \bgroup
  \def\arraystretch{1.1}
  \begin{tabularx}{\linewidth}{XX}
    \toprule
    \multicolumn{1}{c}{\textbf{Feature set description}} & \multicolumn{1}{c}{\textbf{Turn the feature set is computed on}} \\
    \midrule
ASR \& NLU Confidence scores & \hspace{10mm} $t_{n}^u$ \\
Length of user request and system Response & \hspace{10mm} $t_{n}^u$, $t_{n}^s$ \\ 
Time between consecutive user requests & \hspace{10mm} $t_{n}^u$-$t_{n+1}^u$  \\
\textbf{Aggregate - Intent Popularity}  & \hspace{10mm} $t_{n}^u$ \\
\textbf{Un-actionable user request} & \hspace{10mm} $t_{n}^s$ \\
\textbf{Cohesion between system response and user request}  & \hspace{10mm} $t_{n}^u$, $t_{n}^s$ \\
 Length of dialogue  & \hspace{10mm} $t_{0}$-$t_{n}$ \\
\textbf{Topic diversity} & \hspace{10mm} $t_{0}^u$-$t_{n}^u$\\
\textbf{User paraphrasing his/her request} & \hspace{10mm} $t_{n}^u$-$t_{n+1}^u$\\
    \bottomrule
  \end{tabularx}
  \egroup
  \end{adjustbox}
  }
   \caption{Based on Gradient Boosting Regression model output, top 10 feature sets \protect\footnotemark ranked by Importance score. In bold are the new feature sets we introduced.}
   \vspace{-2.5mm}
  \label{top-feature-weights}
\end{table*}
\footnotetext{For confidentiality we are not mentioning the entire list of feature sets used in the model.}

\label{sec:appendix}
\end{document}